\documentclass{article}

\usepackage[final]{nips_2019}

\usepackage[utf8]{inputenc} 
\usepackage[T1]{fontenc}    
\usepackage{hyperref}       
\usepackage{url}            
\usepackage{booktabs}       
\usepackage{amsfonts}       
\usepackage{nicefrac}       
\usepackage{microtype}      
\usepackage{multirow}
\usepackage{color}
\usepackage{amsmath,amssymb} 
\usepackage{pgfplots}
\pgfplotsset{compat=newest}
\usepackage{pdflscape}
\usepackage{graphicx}
\usepackage[export]{adjustbox}
\usepackage{algorithm}
\usepackage{mathtools}
\usepackage{caption,subcaption}
\usepackage{multirow}
\usepackage{bbm}
\usepackage{booktabs}
\usepackage{amssymb}
\usepackage{amsfonts}
\usepackage{dsfont}
\usepackage{mathtools}
\usepackage[inline]{enumitem}
\usepackage{amsthm}
\usepackage{algorithmic}
\usepackage{wrapfig,booktabs}

\newcommand{\dplus}{\mathbin{{+}\mspace{-9mu}{+}}}

\usepackage{array}

\usepackage{wrapfig}

\title{SplitNN-driven Vertical Partitioning}

\author{
  Iker Ceballos \\
  Acuratio\\
  \texttt{iker@acuratio.com} \\
  \And
  Vivek Sharma \\
  MIT/ Harvard Medical School \\
  \texttt{vvsharma@mit.edu} \\
  \And
  Eduardo Mugica \\
  Acuratio \\
  \texttt{eduardo@acuratio.com} \\  
  \And
  Abhishek Singh \\
  MIT \\
  \texttt{abhi24@mit.edu} \\
  \And
  Alberto Roman \\
  Acuratio \\
  \texttt{alberto@acuratio.com} \\
  \And
  Praneeth Vepakomma \\
  MIT \\
  \texttt{vepakom@mit.edu} \\
  \And
  Ramesh Raskar \\
  MIT \\
  \texttt{raskar@mit.edu} \\
}

\begin{document}

\maketitle

\begin{abstract}
In this work, we introduce SplitNN-driven Vertical Partitioning, a configuration of a distributed deep learning method called SplitNN to facilitate learning from vertically distributed features. SplitNN does not share raw data or model details with collaborating institutions. The proposed configuration allows training among institutions holding diverse sources of data without the need of complex encryption algorithms or secure computation protocols. We evaluate several configurations to merge the outputs of the split models, and compare performance and resource efficiency. The method is flexible and allows many different configurations to tackle the specific challenges posed by vertically split datasets.
\end{abstract}

\section{Introduction}

Leading banks and financial services are currently using deep learning algorithms to optimize their processes on targeted tasks, such as approving loans, assessing risk and carrying out credit scores among others. The financial sector generates huge amounts of data daily, and are always in need of better ways to assess risk and detect fraud, and also to utilize the data efficiently. Even if considerable progress has been made in a data-intensive industry, financial services companies face challenges to efficiently adapt to the latest data processing techniques, and there is an increasing pressure to access third party data to improve their operational efficiency. On top of these challenges, there are issues like regulatory compliance costs; competition; legacy infrastructures; security concerns, that prevent financial services companies from effectively use data. Unlocking data silos and uncovering novel sources of data will provide a competitive advantage, and companies in regulated sectors will have higher network effects than its competitors.



Currently, each company works in isolation, where they keep their data private and use that to build their own proprietary models. On the other hand, banks utilize third parties data and resources pulled from several companies as services to build their customized models for their targeted tasks. However, with the recent rise of a new distributed deep learning: SplitNN~\citep{gupta2018distributed} architecture, a new way to process all of these data collaboratively has emerged, without conceding ownership or loosening privacy requirements. Furthermore, using SplitNN~\citep{gupta2018distributed,vepakomma2018split,Sharma2019ExpertMatcherAM} also enables the use of distributed sources of data, which results in improved and generalised robust models. Tapping into this data, which is often private and owned by several companies poses a totally new set of challenges that can be addressed by SplitNN.

Motivated by the above observations, we are interested in the setting where multiple entities (clients) collaborate for a targeted task at hand, under the coordination of a central server or service provider. Each client’s raw data is stored locally and are not exchanged or transferred; instead, focused updates intended for immediate aggregation are used to achieve the learning objective. Usually the data is split horizontally, meaning that each company holds an unique set of features over a non-overlapping set of users. As mentioned earlier, in an industry with huge amounts of data on every client, such as the financial services industry, companies often have enough data of a particular type. In that scenario SplitNN provides more value, since it allows to utilize data from several parties~\citep{Sharma2019ExpertMatcherAM,sharma2019expertmatcher0}. The other variants to SplitNN for distributed private training can also be achieved with Federated Learning~\citep{brendan2017communication,kairouz2019advances}, but unlike SplitNN it is necessary to share the complete model with all the clients. In essence, the goal of our work is to learn a shared model using vertical partitioned data coming from several sources while preserving data privacy.

\section{Related Work} 
In this section, we share related works on several techniques for vertically partitioned machine learning. We categorize these works under the following categories.  
\begin{enumerate}
    \item \textbf{Vertically partitioned linear and logistic regression} The work in \citep{gascon2016secure} proposes a multi-party computation (MPC) scheme based on garbled circuits for secure linear regression in the vertically partitioned setting. The works in \citep{yang2019quasi,yang2019parallel} provide schemes for secure vertically partitioned logistic regression based on homomorphic encryption.
    \item \textbf{Vertically partitioned decision trees} The works in \citep{gascon2016secure, vaidya2008privacy, vaidya2004privacy, kourtellis2016vht, cheng2019secureboost} share approaches for vertically partitioned learning with decision trees, gradient boosted decision trees and Hoeffding trees. 
    \item \textbf{Vertically partitioned SVM} 
    The work in
    \citep{shen2019secure} shows a threshold Paillier and blockchain based secure approach for using support vector machines on vertically partitioned data. Similarly, the work  in \citep{stolpe2013anomaly} shows a simpler approach of using core vector machines for anomaly detection using vertically partitioned data.
    \item \textbf{Vertical federated learning} Learning with vertically partitioned data in the context of federated learning, a popular distributed deep learning paradigm was studied in \citep{nock2018entity,liu2019communication,yang2020communication,Wagh2018SecureNNEA,vepakomma2018split,Sharma2019ExpertMatcherAM,sharma2019expertmatcher0}. Conventional solutions in this setting make use of expensive cryptographic schemes such as Homomorphic Encryption and Multi-Party Computation and thus face critical performance challenges and communication overhead. SecureNN~\citep{Wagh2018SecureNNEA} was proposed in 2018 achieved great success in reducing the communication by over 8 times and in eliminating the requirement to use conventional cost intensive oblivious transfer protocols. \newline

Other lines of work try to avoid these challenges~\citep{yang2020communication} following the same design principles as~\citep{brendan2017communication}, and they propose Federated Stochastic Block Coordinate Descent (Fed-BCD). They show that applying classical Block Coordinate Descent to the FL setting can significantly reduce the communication cost. In their setting they reduce the amount of communication by updating the model fewer times with richer local updates. This approach maximizes the information sent in each update, since having hundreds of clients means each communication round is very expensive. Further, performing local updates on the clients requires sharing the labels which is not always feasible.

Motivated by these observations, \citep{vepakomma2018split} proposed an approach called split learning in which a smaller fraction of the model is present on each client-network and just the output of these models is shared with the server in every iteration. This results in smaller but more frequent updates and it helps reduce communication and computational overhead on the clients.

\end{enumerate}

\section{Vertical SplitNN} \label{sec:method}
\begin{figure*}[t]
\centering
{\includegraphics[width=0.99\columnwidth]{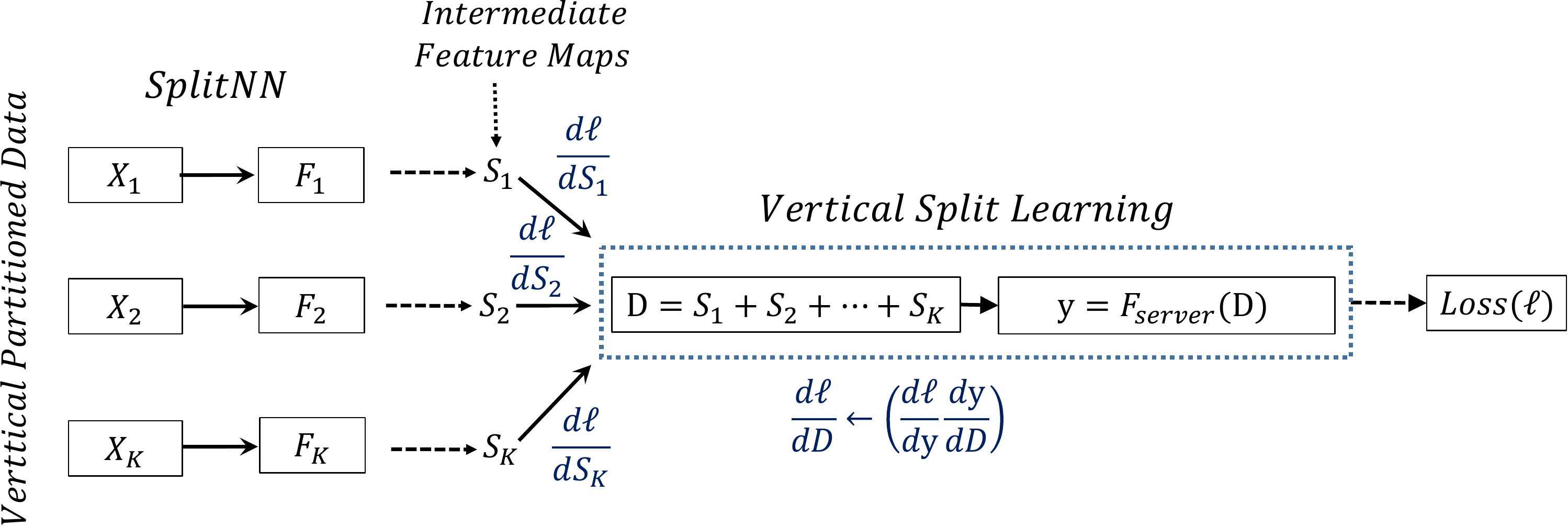}} 
\caption{Vertical SplitNN architecture: Each client computes a fixed portion of the computation graph and passes it to the server which computes the rest and performs back-propagation and returns back the jacobians to the client which can perform their respective back-propagation.}
\label{fig:diagram}
\end{figure*}

The overall goal of our work is to learn a shared model while preserving data privacy. To this end, we propose to train partial neural networks~(NN) on each client and then aggregate all of their outputs before feeding them to the last stage of the combined model on the server-side, as seen in Figure~\ref{fig:diagram}. We are inspired from SplitNN~\citep{vepakomma2018split}. In particular, we extend SplitNN architecture to use all of the partial clients-networks on each iteration instead of using them sequentially. We employ five pooling mechanisms to aggregate the outputs of the partial networks via element-wise average, element-wise maximum, element-wise sum, element-wise multiplication and concatenation.

Among all the aggregation mechanisms, concatenation is the simplest approach and is the closest to training a single network with all of the input features. However, this method requires having the intermediate outputs of all the networks on every iteration, so it is not robust to stragglers. Element-wise sum and average pooling are very close to each other, the main limitation of both of these aggregation methods is that all the networks need to have compatible shapes so that their outputs can be combined together. On the other hand, these methods allow the use of a secure aggregation protocol~\citep{bonawitz2016practical}, which can enhance the privacy and security of the algorithm. Element-wise max pooling also requires all the networks to have compatible output shapes, in this case we pick the activations with the maximum value for each neuron and discard the rest. All these setups require communication on every iteration since they are jointly optimized by back-propagating the error from the main network to the smaller ones. 
One can readily employ other encoding methods like Compact Bilinear Pooling~\citep{cbp,cbp0}, Temporal Compact Bilinear Pooling~\citep{tcbp}, NetVLAD~\citep{netvlad} instead of the pooling mechanisms for a more robust representation learning.

\textbf{Implementation:} Split Learning was defined by~\citep{gupta2018distributed}. 
We utilize SplitNN~\citep{gupta2018distributed} architecture as a baseline architecture for distributed private training.

In our method each partial clients-network encodes its data into a different space and then transmits it to train a shared deep servers-network. A deep neural network can be defined as a function $F$, describable as a sequence of layers $\{L_0, L_1, ...L_N\}$. For a given input $X$, the output of this function is given by $F(X)$ which is computed by sequential application of layers, given as:
\begin{equation*}
    F(X) \leftarrow L_N(L_{N-1}...(L_0(X)))
\end{equation*}

Gradients can be backpropagated over each layer to generate gradients of previous layers and to update the current layer. We will use $L^T_i(gradient)$ to denote the process of backpropagation over one layer and $F^T(gradient)$ to denote backpropagation over the entire neural network. Similar to forward propagation, backpropagation on the entire neural network is comprised of sequential backward passes, given as:
\begin{equation*}
F^T(gradient) \leftarrow L^T_1(L^T_2...(L^T_N(gradient)))
\end{equation*}

The process of sequential computation and transmission followed by computation of remaining layers is functionally identical to application of all layers at once. Similarly  because  of  the  chain  rule in differentiation, backpropagating $F^T(gradients)$  is functionally identical to sequential application of $F^T_a(F^T_b(gradients))$.

When extending this for multiple concurrent clients as in the SplitNN-driven vertical partitioning case, the backpropagated error will be split and each client-network will be passed the corresponding gradients. Let's take concatenation as an example. The full forward pass will be a concatenation of the forward passes coming from each client-network $\{F_1 \dplus F_2 \dplus \ldots F_k\}$, $k \in \{1,\ldots,K\}$ where $K$ is the maximum number of clients. In the same way, the gradients on the concatenation layer will be $\{L_1 \dplus L_2 \dplus \ldots L_k\}$. Thus, we only need to split this gradients before passing them to the upstream clients.

We evaluate our proposed method on three popular financial datasets, namely Bank Marketing~\citep{marketing}, Give me Some Credit~\cite{credit} and Financial PhraseBank~\citep{phrasebank}. We focus on financial datasets because of the special relevance of vertically partitioned data which is widespread in the industry. At the moment, financial institutions share plain and anonymized data with third parties for critical applications, but this is not the best solution due to multiple reasons: little amount of privacy offered by the anonymization scheme, no control on the usage, inability to audit the usage and marginal returns on the value of data is depleted with each partnership. Using our proposed split learning scheme, the need to pool all of the data together is obviated, keeping the data sources private, which enables unprecedented collaboration between the sharing parties and the non-rivalry of data~\citep{jones2018nonrivalry} would potentially lead to increasing returns. Use cases in the industry range from multi-party borrowing detection, risk analysis, fraud detection to cross-selling and customer retention.

\section{Experiments} \label{sec:experiments}
We present our evaluation of the proposed method on several different datasets. All of the datasets here are used for the prediction task.

\textbf{Datasets and Implementation Details.}
We test our system experimentally on three financial datasets. The datasets are summarized in Table~\ref{table:stats}. 

\begin{table*}[ht]
\small
\tabcolsep=1.5mm
\begin{center}
\caption{\textbf{Datasets}. ``\#Samples'' denotes the number of samples, ``\#Dim'' denotes the dimensionality of the features and ``\#Classes'' denotes the number of classes. } 
\label{table:stats}
\resizebox{8cm}{!}{
\begin{tabular}{l|ccc}
\toprule
& bankmarketing & givemecredit & phrasebank \\
\midrule
\#Samples & 45k & 30k & 5k \\
\#Dim. & 16 & 25 & 300 \\
\#Classes & 2 & 2 & 3 \\
\bottomrule
\end{tabular}}
\end{center}
\end{table*}

\textit{Bank Marketing}~\citep{bankingdataset} is a dataset related with direct marketing campaigns of a Portuguese banking institution from UCI machine learning repository. We use all the 16 feature dimensions for prediction and distribute them vertically among the clients. The vertical split is done based on the source of the features, with the bank client data in one split and all the social and economic context attributes in the other.

\textit{Give Me Some Credit}~\citep{givemecreditdataset} is a dataset of financial data built for the task of predicting the likelihood of someone experiencing a financial distress in the close future. Once again there is no coherent vertical split for the data so we choose to split the features arbitrarily into two sets.

\textit{Financial PhraseBank}~\citep{DBLP:journals/corr/MaloSTKW13} consists of 4845 english sentences selected randomly from financial news found on LexisNexis database. These sentences then were annotated by 16 people with background in finance and business. The annotators were asked to give labels according to how they think the information in the sentence might affect the mentioned company stock price. We use all the sentences in the dataset and apply GloVe~\citep{glove} based embedding with 300 dimensions for the word embedding. After applying GloVe we treat this embedding space as a feature space and split it arbitrarily in a four vertical splits.

\textbf{Evaluation Metric.} We report accuracy as well as F1 score to account for the class imbalance.

\textbf{Multiple Clients.} Due to the small number of features available, We use only two splits for both \textit{Bank Marketing} and \textit{Give me Credit} datasets. We use \textit{Financial PhraseBank} to analyze the effect of splitting the dataset across a higher number of clients. From practical standpoint, it is worth mentioning that unlike in horizontally distributed datasets, in vertical SplitNN the number of clients are likely to remain small since relevant data about a single user is not usually distributed into too many sources.

\subsection{Comparison with a centralized model}

In Table~\ref{table:singlevsmax}, we compare the results of training a centralized model~$(M)$ with training several split models~$(M1, M2, M3...)$, merging their outputs and using those as input for $M$. We choose max pooling as a merging technique for this comparison since it overall provides the best performance for the studied datasets.

As noted in the results, the Financial Phrasebank dataset is the only one where vertical partitioning and element-wise max pooling results in a drop in performance. This could be due to two reasons - the data we applied vertical partitioning over was obtained after the embedding, splitting the 300 GloVe features into 4 sets, the other reasoning could be due to the underlying semantic nature of a sentence making it a difficult task for vertical partitioned learning from a practical standpoint. For all the other cases, the performance roughly remains same with some marginal improvements when using split learning.

\begin{table*}[ht]
\small
\tabcolsep=1.5mm
\begin{center}
\caption{Comparison of the performance of a single model with access to the full dataset vs a split model with four vertical partitions. We only report results for element-wise max pooling since it's the best performing merging technique.} 
\label{table:singlevsmax}
\resizebox{8cm}{!}{
\begin{tabular}{l|cc|cc}
\toprule
& \multicolumn{2}{c}{Single Model} & \multicolumn{2}{c}{Max Pooling} \\
Dataset & Acc & F1 & Acc & F1 \\
\midrule
Bank Marketing & 0.83 & 0.47 & 0.84 & 0.47 \\
Give Me Credit & 0.80 & 0.34 & 0.81 & 0.35 \\
Financial PhraseBank & 0.78 & 0.78 & 0.76 & 0.76 \\
\bottomrule
\end{tabular}}
\end{center}
\end{table*}

\subsection{Comparison of merging strategies}

In Table~\ref{table:allmerging}, we compare several strategies to merge the outputs of the models trained with the vertically partitioned features. We consider two pooling mechanisms as well as simple combinations such as concatenation, element-wise multiplication and element-wise sum of the outputs.

The simplest strategy is the concatenation, however this requires all outputs from each participating client to be present during the forward pass, which could be infeasible in a real scenario since some of the clients may drop randomly or there might be synchronization issues. Therefore any of the other strategies are preferable because of their aggregation mechanism.

Furthermore, both element-wise average pooling and simple element-wise addition over the inputs can allow us to use a secure aggregation protocol while combining the outputs of the smaller models. Thus, providing an extra layer of security on top of the obfuscation provided by the models themselves and NoPeek~\citep{vepakomma2018no}.

We notice that the performance doesn't suffer huge drops with any of the methods. However, in practice one could choose the average pooling since, it allows to use a secure aggregation protocol as well as compression techniques that can help with stronger privacy and communication overhead respectively.

Figure~\ref{fig:merge} shows the loss and metrics during training for Financial PhraseBank. The centralized training (single model) takes fewer batches to converge, 

\textbf{\begin{figure*}[t]
\makebox[\textwidth][c]{\includegraphics[width=1.3\textwidth]{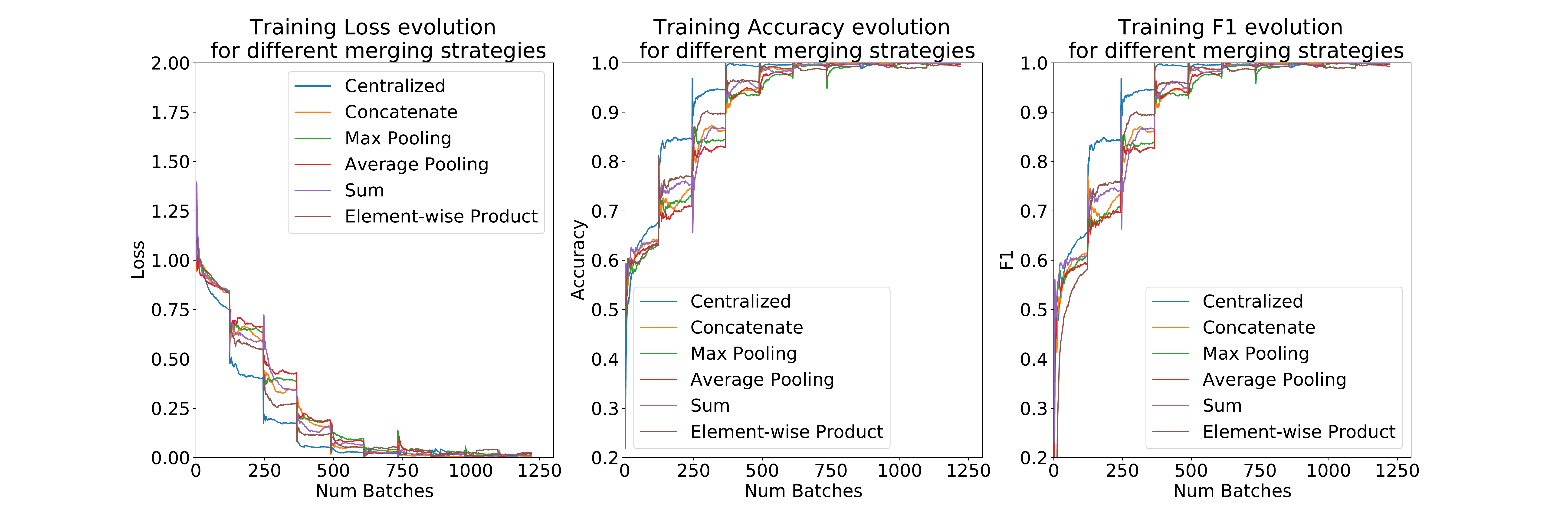}}
\caption{Comparison of several merging strategies for SplitNN-driven vertical partitioning with PhraseBank.}
\label{fig:merge}
\end{figure*}
}

\begin{table*}[ht]
\small
\tabcolsep=1.5mm
\begin{center}
\caption{Comparison of merging/pooling strategies.} 
\label{table:allmerging}
\resizebox{12cm}{!}{
\begin{tabular}{l|cc|cc|cc}
\toprule
& \multicolumn{2}{c}{Financial PhraseBank}  & \multicolumn{2}{c}{Bank Marketing} & \multicolumn{2}{c}{Give me Credit} \\
Merging & Acc & F1 & Acc & F1 & Acc & F1 \\
\midrule
Element-wise Max Pooling & 0.76 & 0.76 & 0.84 & 0.47 & 0.81 & 0.35 \\
Element-wise Average Pooling & 0.77 & 0.77 & 0.83 & 0.46 & 0.82 & 0.36 \\
Concatenation & 0.76 & 0.76 & 0.82 & 0.46 & 0.83 & 0.37 \\
Element-wise Multiplication & 0.72 & 0.72 & 0.82 & 0.46 & 0.80 & 0.34 \\
Element-wise Sum & 0.77 & 0.76 & 0.83 & 0.46 & 0.77 & 0.32 \\
\bottomrule
\end{tabular}}
\end{center}
\end{table*}

\subsection{Clients dropping randomly}

In Table~\ref{table:dropping}, we present the results of dropping some of the clients randomly both during training and testing.

The drop during the training means that the model is trained with the outputs of all models but on each iteration one or more of those outputs is missing. On the other hand, dropping during testing means that the model was trained with the outputs of all the models but for the prediction on the test set the output of some of the models from the client side is missing.

\begin{table*}[ht]
\small
\tabcolsep=1.5mm
\begin{center}
\caption{Comparison of merging strategies when clients drop randomly. We report accuracy for the Financial PhraseBank.} 
\label{table:dropping}
\resizebox{12cm}{!}{
\begin{tabular}{l|ccc|ccc|}
\toprule
& \multicolumn{3}{c}{Training} & \multicolumn{3}{c}{Testing} \\
Merging & Drop 1 & Drop 2 & Drop 3 & Drop 1 & Drop 2 & Drop 3 \\
\midrule
Element-wise Max Pooling & 0.74 & 0.72 & 0.69 & 0.76 & 0.70 & 0.63 \\
Element-wise Average Pooling & 0.75 & 0.72 & 0.69 & 0.74 & 0.71 & 0.65 \\
Element-wise Multiplication & 0.75 & 0.75 & 0.71 & 0.71 & 0.60 & 0.58 \\
Element-wise Sum & 0.74 & 0.73 & 0.70 & 0.74 & 0.70 & 0.64 \\
\bottomrule
\end{tabular}}
\end{center}
\end{table*}

As shown in the Table~\ref{table:allmerging}, in both the cases the performance suffers a significant impact as a consequence of the clients dropping. This is expected, since we are missing the predictive power of several features. When we increase the number of clients that drop at each point the performance hit is even bigger, which is consistent with our hypothesis.

Furthermore, in Figure~\ref{fig:drop} we can see that dropping more than two clients, in a four client setting, even affects the convergence of the model and the loss starts to rise by the end of the training, indicating that optimization is drifting from local/global minima. This performance drop however does not arise if a client drops just on a few iterations but is present for most of the training. This is an interesting starting point for future work since it would be interesting to analyze how to minimize the impact of stragglers with vertical SplitNN.

\textbf{\begin{figure*}[t]
\centering
\makebox[\textwidth][c]{\includegraphics[width=1.3\textwidth]{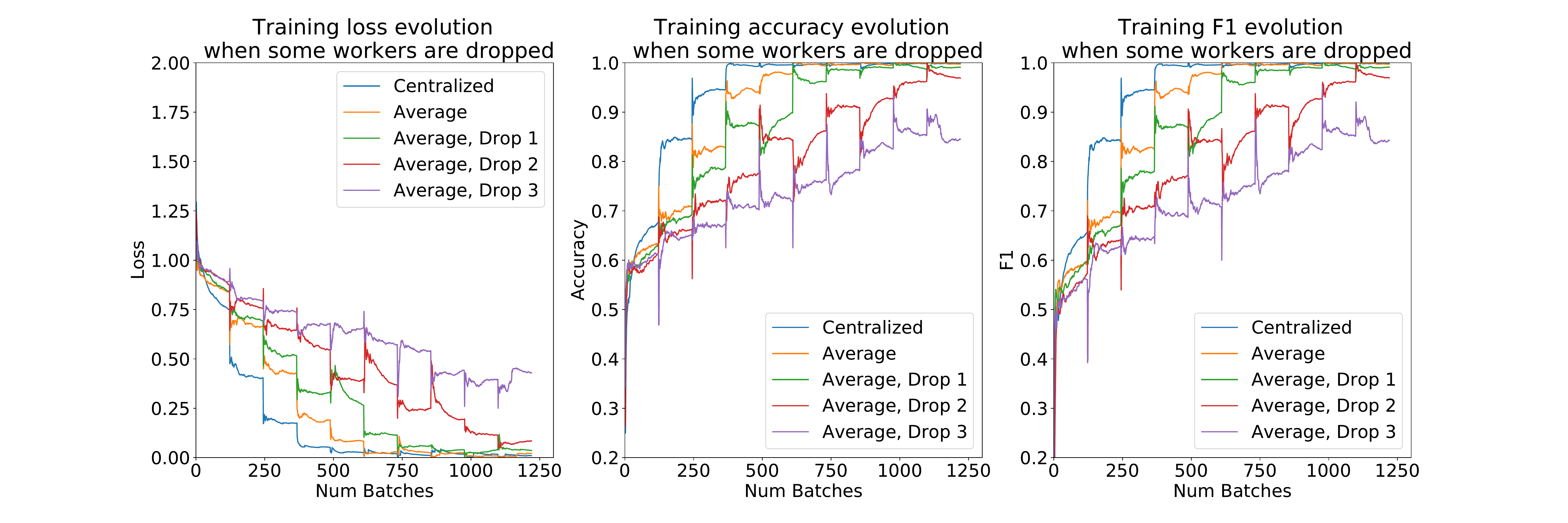}}
\caption{Loss and metrics for PhraseBank dataset while workers drop during training.}
\label{fig:drop}
\end{figure*}
}

\subsection{Measurement of communication and computational costs}

Tests over different datasets were carried out in order to estimate the amount the communications performed in a vertical SplitNN training process. We use the roles defined in \citep{ceballos2020systemdesign} to identify the type of data available for each of the participants. Role 1 only has access to features, role 3 has access to both features and labels and role 0 is just a computation client with no data. Each test was carried out by three clients. One of them with role 1, another one with role 3 and the last one with role 0. Results are shown in Table~\ref{table:communication}.

\begin{table*}[ht]
\small
\tabcolsep=1.5mm
\begin{center}
\caption{Communication costs in initialization, forward pass and backward pass for each studied dataset.} 
\label{table:communication}
\resizebox{14cm}{!}{
\begin{tabular}{l|ccc|ccc|ccc}
\toprule
Dataset & \multicolumn{3}{c}{Financial PhraseBank} & \multicolumn{3}{c}{Bank Marketing} & \multicolumn{3}{c}{Give me Credit}\\
Role & 1 & 3 & 0 & 1 & 3 & 0 & 1 & 3 & 0 \\
\midrule
Total sent per epoch (MB) & 488 & 490 & 977 & 2,560 & 3,840 & 7,680 & 4,800 & 7,200 & 14,400 \\
\midrule
Total received per epoch (KB) & 488 & 490 & 977 & 2,560 & 5,120 & 6,400 & 4,800 & 9,600 & 12,000 \\
\bottomrule
\end{tabular}}
\end{center}
\end{table*}

The communication cost computed for this table is based on the division of task according to different roles. Once the training starts for each batch, workers with roles 1 and 3 send the output of their next-to-last layer to role 0 worker, which performs its forward pass and send the output of its next-to-last layer to worker with role 3 to compute the loss.

Similarly once the loss is computed, role 3 worker will send back to role 0 worker the error at the output of the shared layer so that it can continue back propagating it. Finally this role 0 worker will send back to its corresponding worker with role 1 or 3 the error at the output of each corresponding shared layer. 

\begin{table*}[ht]
\small
\tabcolsep=1.5mm
\begin{center}
\caption{Measurements of the computational costs} 
\label{table:computation}
\resizebox{12cm}{!}{
\begin{tabular}{l|ccc}
\toprule
Dataset & Financial PhraseBank & Bank Marketing & Give me Credit \\
\midrule
Number of parameters of the NN & 3,907,059 & 745 & 457 \\
FLOP/sample & 33,667 & 4,041 & 741 \\
us/batch (batch size=32) & 26,037 & 911 & 793 \\
MFLOPS (batch size=32) & 41.377 & 141.945 & 29.902 \\
us/batch (batch size=128) & 97,871 & 1,114 & 1,107 \\
MFLOPS (batch size=128) & 44.031 & 464.316 & 85.680 \\
\bottomrule
\end{tabular}}
\end{center}
\end{table*}

The communication size in a vertical SplitNN architecture is dependent on the size of the output at the endpoints layer. 
The computational cost however is dependent on the architecture and the size of the input feature vector at each layer. For widely used architectures, everything remains same here in comparison with traditional deep learning except the size of the feature vector of the first layer on the central server.

Bearing this in mind, in conjunction with performance trade-offs between different merging strategies, it is extremely important to know the details and the limitations of the specific use case, in order to be able to propose the best training strategy. The neural network architecture splitting scheme between the workers, or the adjustment of the hyper-parameters can greatly change the speed and therefore the efficiency of the training process.

Thus, we find that in the training processes where the bottleneck is on the communication side, most of the training process should be done in workers with roles 1 and 3 so that the outputs of their networks are already as small as possible. 
On the other hand when the bottleneck is the computational cost, workers with roles 1 and 3 should have the minimum amount of layers to assure the data is kept private, and the core of the model should be in a role 0 worker with higher computational capacity. As shown in Table~\ref{table:computation}, other techniques as adjusting the batch size could be highly convenient in some cases to speed up the training processes.

An interesting line of research for the future, would be to study the effect on the convergence of compression techniques such as STC~\citep{sattler2019stcfederated} or Random Rotation Matrix~\citep{konecny2017commefficiency} as well as privacy preserving techniques such as Secure Aggregation Protocol~\citep{bonawitz2016practical} or minimizing Distance Correlation~\citep{vepakomma2019reducing}, as well as their effect on the computational cost.

\section{Conclusion} \label{sec:conclusion}
In this paper, we proposed split learning for vertically partitioned data and further addressed the specific challenges arising in this scenario.
We have shown that  the proposed methods to merge the outputs of the split networks result in a shared model that performs on par with the centralized model. Max-pooling is the best overall, however we believe the small drop in performance shown with average-pooling is acceptable considering that it allows the use of a secure aggregation protocol. We believe our approach to train models with vertically partitioned data provides a way which is better suited to its specific challenges, which are different from those arising with horizontally partitioned data.



\bibliographystyle{authordate1} 
\bibliography{main}

\end{document}